%% file: IEEE-conference-template-062824.tex
\documentclass[conference]{IEEEtran}
\IEEEoverridecommandlockouts

\usepackage{cite}
\usepackage{amsmath,amssymb,amsfonts}
\usepackage{algorithmic}
\usepackage{graphicx}
\usepackage{textcomp}
\usepackage{xcolor}

\usepackage{algorithm}
\usepackage{algorithmic}
\usepackage{times}
\usepackage{soul}
\usepackage{url}
\usepackage[utf8]{inputenc}
\usepackage{graphicx}
\usepackage{amsmath}
\usepackage{amsthm}
\usepackage{booktabs}
\usepackage[switch]{lineno}

\usepackage{subfigure}
\usepackage{xspace}
\newcommand{\ourmethod}{FedC$^2$I\xspace}
\usepackage{pifont} 
\theoremstyle{definition}
\newtheorem{definition}{Definition}[section]
\usepackage{amssymb}
\usepackage{makecell}
\usepackage{multirow}
\usepackage{color}
\usepackage{xcolor}

\def\BibTeX{{\rm B\kern-.05em{\sc i\kern-.025em b}\kern-.08em
    T\kern-.1667em\lower.7ex\hbox{E}\kern-.125emX}}
\begin{document}

\title{Influence-Oriented Personalized Federated Learning}

\author{
\IEEEauthorblockN{Yue Tan\IEEEauthorrefmark{1}\IEEEauthorrefmark{2}, 
Guodong Long\IEEEauthorrefmark{2}, 
Jing Jiang\IEEEauthorrefmark{2}, 
and Chengqi Zhang\IEEEauthorrefmark{3}}
\IEEEauthorblockA{
\IEEEauthorrefmark{1}Griffith University, 
\IEEEauthorrefmark{2}University of Technology Sydney, 
\IEEEauthorrefmark{3}Hong Kong Polytechnic University \\
yue.tan@griffith.edu.au, guodong.long@uts.edu.au, jing.jiang@uts.edu.au, chengqi.zhang@polyu.edu.hk
}
}

\maketitle

\begin{abstract}
Federated learning (FL) is a machine learning paradigm where clients with different behaviors and preferences can learn collaboratively without compromising data privacy. 
Typical FL methods often rely on fixed weighting for parameter aggregation, thereby neglecting the mutual influence among clients.
In practice, clients with similar preferences or backgrounds may provide more useful knowledge to each other, which can be leveraged to improve local performance. 
However, how to quantify such cross-client influence and how to exploit it for personalized aggregation remain underexplored. 
To address this gap, we propose an influence-oriented \underline{Fed}erated learning framework which quantitatively measures \underline{C}lient-level and \underline{C}lass-level \underline{I}nfluence to realize adaptive parameter aggregation for each client (\underline{\ourmethod} for short). 
Our core idea is to explicitly model the inter-client influence within an FL system via the well-crafted influence vector and influence matrix.
Specifically, \ourmethod incorporate influence vectors to quantify client-level influence, enables clients to selectively acquire knowledge from others, and guides the aggregation of feature representation layers.
Meanwhile, the influence matrix captures class-level influence in a more fine-grained manner to achieve personalized classifier aggregation. 
We evaluate the performance of \ourmethod against existing federated learning methods under non-IID settings, and the results demonstrate the superiority of our method in terms of effectiveness, robustness, and interpretability. 
\end{abstract}


\input{icdm26_content}



\bibliographystyle{IEEEtran}
\bibliography{icdm26}

\end{document}

%% file: icdm26_content.tex
\section{Introduction}
Federated learning (FL) is a promising machine learning paradigm where multiple clients with diverse behaviors and preferences can collaboratively train models without sharing their private data~\cite{mcmahan2016communication}. A core challenge in FL is the statistical heterogeneity, also known as the non-IID problem, \textit{i.e.}, clients may own data samples from different domains with different feature attributes. The data distribution shift over multiple clients can prevent the globally learned model from achieving high performance on local data distribution. The vanilla FL method, namely FedAvg, significantly suffers from performance degradation due to the widely existing non-IID data in real-world scenarios. Existing works aim to alleviate this by adding regularization terms~\cite{wang2020tackling,zhang2021parameterized}, local model decoupling~\cite{luo2022disentangled,chen2022on,collins2021exploiting}, etc. However, most of these works still follow the traditional model aggregation rule where each client is assigned a fixed weight (mostly proportional to its dataset size) corresponding to its local models without considering the underlying mutual effect between any two clients. 

To better exploit client relationships during aggregation, several existing studies uses loss or gradient similarity-based aggregation schemes to reweight each client~\cite{ribero2020communication,balakrishnan2021diverse}. Despite their effectiveness in standard federated learning scenarios, these studies aim to learn a robust and general model by selecting a set of the most representative and informative clients rather than constructing a personalized reweighting scheme for each client to improve their local performance. 
{As a result, in these methods, the less informative clients tend to stay unexplored during the federated training process, degrading the local performance at these clients~\cite{balakrishnan2021diverse}. }
Moreover, the model aggregation procedure in the above approaches is still conducted at the server side, hindering a customized aggregation for each client. This motivates us to ask: \textbf{\textit{how can each client selectively acquire useful knowledge from others according to its own data characteristics and learning needs?}}

\begin{figure}[t!]
    \centering
    \begin{center}
    \subfigure[]{
    	\includegraphics[height=4.8cm]{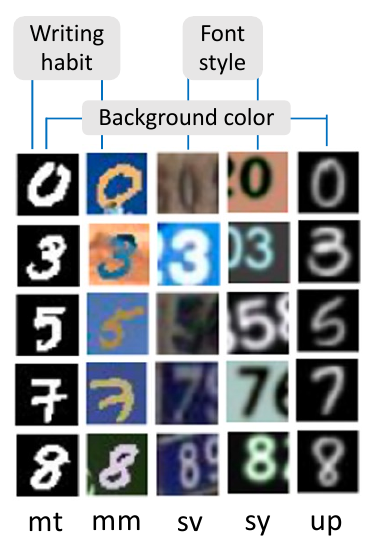}
    	\label{fig:illu-a}
    }
    \subfigure[]{
    	\includegraphics[height=4.8cm]{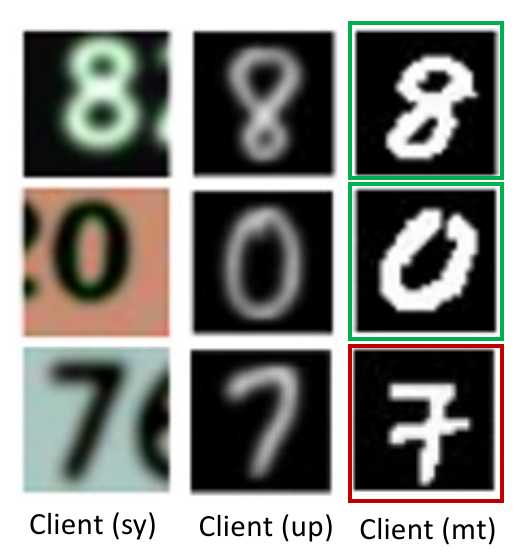}
    	\label{fig:illu-b}
    }
    \end{center}
    \caption{(a) Five clients owning digit images from various domains. Some share the same background color, font style, and/or hand-writing habit, forming an underlying correlation among clients. (b) Different understandings of digit ``7" held by different clients. }
    \label{fig:illustration}
    \vspace{-4mm}
\end{figure}

A key step to answering this question lies in characterizing the \textbf{cross-client influence}. {As humans, we often enhance our intelligence by learning from others in collaborative environments, and we are especially influenced by those who share similar preferences or backgrounds. Inspired by this, a natural solution is to develop an influence-oriented personalized federated learning framework which allows each client to improve its local model by leveraging the influence of other models. Specifically, influence is measured by the consistency between a client's local model and those of its neighbors, which can indicate how much knowledge a client is expected to gain from others during aggregation. 
Based on this idea, a follow-up question naturally arises: \textbf{\textit{what kinds of influence should be considered in personalized federated learning?}}

Through analyzing the characteristics of real-world heterogeneous data, we identify two types of representative influences. 
The first type is \textbf{client-level influence}, which captures shared class-related characteristics 
among samples of the same category across different clients. 
As shown in Fig.~\ref{fig:illu-a}, there are five clients owning digit images from five different domains. Some of them explicitly share informative prior knowledge, \textit{e.g.}, the background color, font style, and/or hand-writing habit, suggesting the underlying correlation among clients with heterogeneous data. Explicitly capturing the client-level influence can help each client identify beneficial collaborators and selectively acquire useful knowledge from them during personalized aggregation. 
Apart from client-level influence, clients may have similar or different understandings toward a specific class, highlighting the necessity of modeling \textbf{class-level influence}.  {Taking} digit classification as an example in Fig.~\ref{fig:illu-b}, most clients share the same understanding toward the digit ``0" and digit ``8". In contrast, as for digit ``7", the client in the right column is prone to capture the detail of digit ``7" with a slash through it, while others hold opposite opinions. Modeling the class-level influence can help each client distinguish beneficial and harmful knowledge 
for each specific class, which is complementary to client-level influence.

Motivated by the above insight and observation, we propose an influence-oriented \textbf{Fed}erated learning framework where both {\textbf{C}lient-level} influence and {\textbf{C}lass-level} \textbf{I}nfluence are explicitly quantified and utilized to guide the parameter aggregation (\textbf{\ourmethod} for abbraviation). 
Specifically, to measure the influence of a specific client on another one during parameter sharing, we follow the leave-one-out principle and observe how the local performance changes if a client does not contribute to the aggregated model. We introduce the influence vector with each element specifying the influence brought by a client in the FL system. By utilizing the influence vector to aggregate model parameters of the feature representation layers, underlying client-wise correlative knowledge {can be} shared more efficiently. 
To further identify the class-level disagreement and address it during the aggregation procedure, we introduce an influence matrix to measure class-level influence{,} which {provides} a more fine-grained metric to investigate how a specific class in a client makes its contribution to the current client. 
The influence matrix is capable of capturing class-level influence and leading to a better parameter aggregation of classifiers. 
By measuring the client-level and class-level influence, clients are capable of aggregating the feature representation layers and the classifier with a unique weight allocation among all participating clients. Therefore, the knowledge conveyed by the model parameters can be acquired by clients in an efficient and personalized way. In summary, the main contributions are summarized as follows:
\begin{itemize}
    \item We take the first step toward exploring the client-level and class-level influence within an FL system, which considerably improves the original parameter aggregation mechanism.
    \item We propose an influence-oriented federated learning framework, namely, \ourmethod, which provides a concrete and feasible solution to quantify the influence and enable clients to personalize the model aggregation procedure.
    \item We carry out extensive experiments on benchmark datasets to show the superiority of \ourmethod compared with baselines and verify the effect of key components in \ourmethod.
\end{itemize}

\section{Related Work}
\subsection{Federated Learning}
Federated learning (FL) enables multiple clients to collaboratively train a global model and/or multiple local models in a distributed manner without sharing their private data~\cite{zhang2021survey,li2020federated,mothukuri2021survey,zhao2026fedcigar}. One core challenge in FL is the statistical heterogeneity issue across clients, also known as non-IID problem~\cite{zhao2018federated,tan2023taming}, where local datasets of clients may have heterogeneous distributions in label and/or feature spaces~\cite{kairouz2021advances}, degrading the local performance and resulting in unstable and slow convergence~\cite{karimireddy2020scaffold}.

Various works are proposed to deal with the \textit{heterogeneous problems} in FL. 
There are recent studies proposing to use clustering-based techniques to improve the vanilla parameter aggregation schemes~\cite{sattler2019clustered,ghosh2020efficient,long2022multi}. In these methods, local models are usually clustered into multiple groups according to different clustering strategies. \cite{jiang2019improving} and \cite{fallah2020personalized} apply meta-learning in FL to obtain a better-initialized model, which can be adapted to various clients by several local training steps. Other studies solve the heterogeneous problems via model decoupling~\cite{li2020fedbn,shen2022cd2,arivazhagan2019federated,qian2026dynhd}, representation learning~\cite{oh2022fedbabu,liang2020think,chen2025multi,li2026relationalAD}, knowledge distillation~\cite{li2019fedmd,shen2022cd2}, etc. Most of the above methods still follow the traditional parameter aggregation pattern where each client is assigned a fixed weighting coefficient, mostly in proportion to their data sizes. In this case, how a client is influenced by others and how it contributes to others is not explicitly measured or discussed. 

There is also a branch of work that investigates \textit{client reweighting} in FL. Most of them introduce centralized client reweighting schemes where the server employs different aggregation weights to clients according to their local performance~\cite{tang2022fedcor,cheng2024fedgcr,wu2026fedram}, model update~\cite{wan2022shielding,ribero2020communication}, and consensus on a public dataset~\cite{zhang2021parameterized,feng2021kd3a}. Some of these works aim to increase the rate of convergence when there are numerous clients in an FL system. The others focus on learning a robust global model by reweighting clients at the central server.

\begin{figure*}[t!]
    \centering
    \setlength{\abovecaptionskip}{-0.0cm}
    \setlength{\belowcaptionskip}{-0.4cm}
    \begin{center}
    \subfigure[An overview of \ourmethod]{
    		\includegraphics[height=5.2cm]{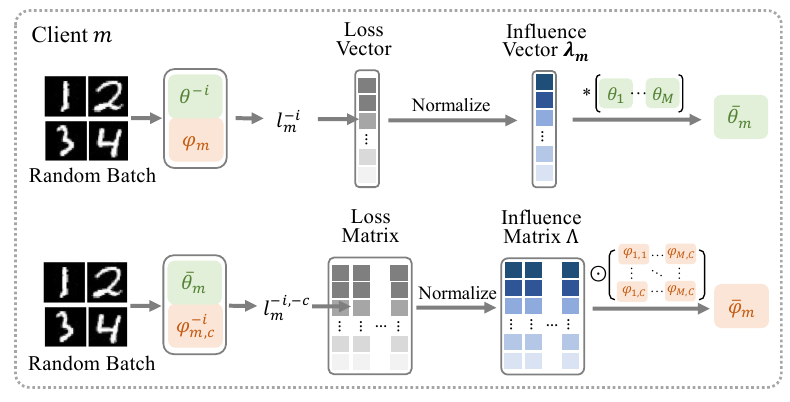}
    	\label{fig:local_gnn}
    }	
    \hspace{0.2cm}
    \subfigure[Client update procedure]{
    		\includegraphics[height=5.2cm]{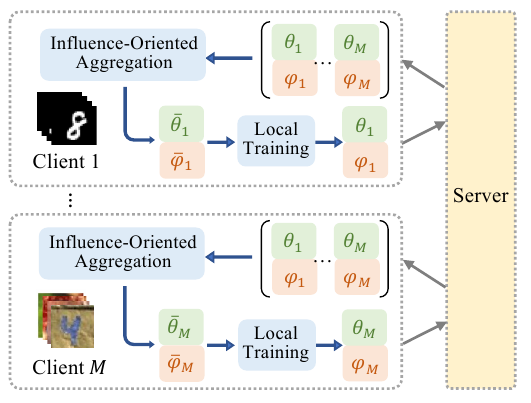}
    	\label{fig:client_update}
    }
    \end{center}
    \caption{(a) An overview of the influence-oriented aggregation procedure in our proposed \ourmethod. Green boxes and orange boxes correspond to the feature representation layers and classifiers, respectively. 
    The loss vector is comprised of loss values. Blue boxes correspond to the models that are trained locally and personalized for each client. Orange boxes correspond to the models aggregated at the server, with knowledge shared across clients. (b) An illustration of the data distribution and model aggregation scheme. }
    \label{fig:framework}
    \vspace{-4mm}
\end{figure*}

\subsection{Influence-Oriented Deep Learning}
There is a branch of works aiming at identifying the influence of training samples in the context of deep learning. In \cite{koh2017understanding}, the authors aim to use the results of leave-one-out retraining to identify the contribution of training points for a given prediction. It focuses on approximating the contribution via influence function in a simple but efficient way. In \cite{jia2019towards} and \cite{ghorbani2019data}, the influence of training samples is identified based on the accuracy of the model. \cite{jia2019towards} utilizes the Shapley value originated in cooperative game theory to evaluate the profit brought by each data. \cite{ghorbani2019data} further develops Monte Carlo and gradient-based methods to efficiently approximate the Shapley values in some real-world settings. 
\cite{yeh2018representer} presents a deeper understanding by decomposing the neural network into multiple activations of training samples and associating it with the influence of training points. \cite{pruthi2020estimating} proposes a general and simple method to compute the influence of a training example on the prediction and applies it to various machine learning models. 

To facilitate the FL system by influence-oriented techniques, \cite{wang2019measure} and \cite{xue2021toward} propose to measure the individual contribution during the collaborative training procedure. \cite{wang2019measure} calculate the instance-level influence by removing it during the training process and use the result to calculate the contribution of a client. \cite{xue2021toward} quantifies the influence over the model parameters and designs an estimator to improve robustness and efficiency. {Nevertheless, both of them use influence measurement for client selection from the perspective of global optimization, while our work focuses on parameter aggregation in a personalized manner. } 

\section{Problem Formulation}
In this section, we provide the basic formulation of the federated learning problem. We consider an FL system with $M$ clients and a central server, {where the $m$-th ($m \in [1, \cdots, M]$)} client owns a private dataset $\mathcal{D}_m$. In this work, we consider the following  optimization objective for the FL framework,
\begin{equation} \label{eq:fl}
\min _{\left\{\mathbf{w}_{1},\mathbf{w}_{2}, \cdots, \mathbf{w}_{M}\right\}} \frac{1}{M} \sum_{m=1}^{M} \lambda_m \mathcal{L}_m\left(\mathbf{w}_{m} ; \mathcal{D}_{m}\right),
\end{equation}
\noindent 
{
where $\mathbf{w}_m$ and $\mathcal{L}_m$ are the model parameters and loss function of the $m$-th client, respectively. ${\lambda}_m$ is the weight of client $m$ when performing parameter aggregation. Here ${\lambda}_m$ is often proportional to the size of local dataset, \textit{i.e.}, ${\lambda}_m = \frac{|\mathcal{D}_m|}{N}$ where $N$ is the total number of instances over all clients.
}

Usually, for traditional classification tasks, the local model of the $m$-th client, parameterized by $\mathbf{w}_m$, can be decoupled into two parts{: (1)} the feature representation layers parameterized by $\mathbf{\theta}_m$, and {(2)} the classifier parameterized by $\mathbf{\varphi}_m$. Suppose there are $C$ classes, $\mathbf{\varphi}_m$ can be further written as 
\begin{equation} \label{eq:wc}
\mathbf{\varphi}_m=\left[\mathbf{\varphi}_m^{1}, \mathbf{\varphi}_m^{2}, \cdots, \mathbf{\varphi}_m^{C}\right],
\end{equation}
\noindent where $\mathbf{\varphi}_m^{c}$ is a weight vector related to the $c$-th class in the linear classifier. This decomposition is particularly suitable for personalized FL under feature heterogeneity. The representation layers $\theta_m$ encode transferable feature patterns, such as shape, texture, style, or domain-specific visual attributes. These patterns may be partially shared among clients even when their data distributions are non-IID. In contrast, the classifier $\phi_m$ maps the learned representations to class predictions and is more directly related to the local label semantics and class-wise decision boundaries. Therefore, aggregating the whole model with a single set of weights may be suboptimal, since the optimal sharing pattern for representation learning may differ from that for classification.

Based on this observation, FedC$^2$I performs decoupled aggregation for $\theta_m$ and $\phi_m$. The representation layers are aggregated according to client-level influence, which reflects how much each client contributes to the target client's overall representation learning. The classifier is aggregated in a class-wise manner according to class-level influence, allowing each target client to selectively absorb class-specific knowledge from different clients. This design provides a more flexible aggregation mechanism than standard model averaging.

To learn the optimal model parameters for each client and surpass the pure local training performance, a variety of FL approaches propose different parameter aggregation and/or local training strategies. Vanilla FL computes the weighted average of all model parameters as
\begin{equation} \label{eq:para_agg}
\overline{\mathbf{w}} \leftarrow \sum_{m=1}^{M} \frac{\left|\mathcal{D}_{m}\right|}{N} \mathbf{w}_{m}.
\end{equation}

For classification tasks, each local model can be decoupled into two parts, \textit{i.e.}, feature representation and classifier. The model parameters can be partially shared for different purposes. For example, in FedRep~\cite{collins2021exploiting}, only the feature representation is shared across clients. In FedRoD~\cite{chen2022on}, in addition to the feature representation, one branch of the classifier is shared, and the other branch of that is locally trained. 

\section{FedC$^{2}$I: An Influence-Oriented Federated Learning Framework}

\subsection{An Overview of \ourmethod} \label{subsec:overview}

{To capture and exploit} the mutual effect among different clients in a personalized FL system, the core idea of \ourmethod is to quantify the \textit{client-level} and \textit{class-level} influence with well-crafted measurements and then execute local parameter aggregation with the guidance of these measurements. 
To measure the influence reasonably, we leverage the leave-one-out principle to {estimate} the client-wise and class-wise contributions {effectively}. 

A brief pipeline of \ourmethod is demonstrated in Fig.~\ref{fig:framework}. 
To represent how other clients influence the $m$-th client, we introduce an influence vector computed from the loss values from multiple leave-one-out models. 
Meanwhile, to identify the class-level disagreement, we construct an influence matrix that quantifies the class-level correlations from a more fine-grained perspective. 
At the beginning of local training, the client-level and class-level influence measurements serve as the weights for parameter aggregation of the feature representation layers and classifier, respectively. In this case, clients are capable of learning a more powerful local model by assimilating cross-domain knowledge in a personalized way. x{The following subsections respectively introduce these crucial designs in detail.}

\subsection{Motivation of Influence-Oriented Aggregation}
The key intuition behind FedC$^2$I is that the usefulness of a client should be evaluated from the perspective of each target client, rather than by a globally shared criterion. In heterogeneous FL, two clients may exhibit similar local structures or domain attributes even if their data distributions are not identical. For instance, two digit-domain clients may share similar writing styles or background patterns, while differing from other clients in color or image quality. In such cases, enforcing the same aggregation weights for all clients may either dilute useful knowledge from highly related clients or introduce negative transfer from irrelevant clients. Therefore, a desirable aggregation mechanism should be both \textit{client-adaptive} and \textit{influence-aware}.

A direct way to estimate such relevance is to compare local data distributions across clients. However, this is incompatible with the privacy-preserving nature of FL, since local data and distributional statistics are not expected to be shared. FedC$^2$I instead estimates inter-client influence through model behavior. Specifically, if removing the model parameters of a certain client leads to a larger loss on the target client's local batch, then this client is considered more influential to the target client. This leave-one-out design provides a simple yet effective way to quantify the contribution of each client without accessing its raw data.

Moreover, influence should not be restricted to the whole-model level. In classification tasks, the feature representation layers and the classifier play different roles. The representation layers capture domain-level and feature-level knowledge, while the classifier contains class-discriminative decision boundaries. Under feature heterogeneity, clients may share useful representations at the domain level but still disagree on the decision pattern of certain classes. Therefore, FedC$^2$I measures influence from two complementary levels: client-level influence for representation aggregation and class-level influence for classifier aggregation.

\subsection{Client-Level Influence Measure} \label{subsec:client_level}
{In an FL system, sharing the knowledge among clients with similar data patterns can mutually boost the local performance, while sharing between clients with totally different data patterns tends to be less helpful to each other. }
Most existing FL methods neither explicitly model this property nor leverage this property to {during parameter aggregation}. 
{If we can quantitatively measure how a local model contributes to another and consider the contribution when aggregating parameters, the local performance can be further improved potentially.} 
{To achieve this, a} direct solution is to analyze the local data located at the client side or transmit sensitive data-related information. However, it will induce huge privacy concerns and make the FL system unreliable~\cite{lyu2022privacy}. 
Therefore, it remains an open problem to figure out how a client contributes to others via the model parameters.
Inspired by recent studies that measure the influence of training samples on prediction results~\cite{koh2017understanding}, we propose to measure the client-level influence by the leave-one-out principle before the model aggregation procedure. 

{To understand how client $m$ is influenced by client $i$ (where $m, i \in [M]$ and $m \neq i$), } 
we start by removing the contribution of client $i$ when performing model aggregation. 
Without the model parameter of client $i$, the parameter aggregation is only conducted over $M-1$ clients rather than $M$ clients. 
In this way, the variation on local performance, \textit{i.e.}, the value of local training loss, can suggest whether client $i$ has a positive impact on client $m$ or not and can further quantify how much impact is brought by client $i$. 

x{Formally, at} client $m$, we use $\{\theta^{-i}, \varphi_{m}\}$ to denote the aggregated model parameter after removing the contribution of client $i$. x{Concretely,} $\{\theta^{-i}, \varphi_{m}\}$ is obtained by
\begin{equation}
\theta^{-i}=\frac{1}{M-1} \sum_{m \in [M], m \neq i} {\theta}_{m}.
\end{equation}
\noindent x{Then, we compute the loss of $\{\theta^{-i}, \varphi_{m}\}$ on a random batch sampled from $\mathcal{D}_m$, where the loss value is denoted as $l_m^{-i}$. } 
Here, $l_m^{-i}$ is a metric that is not only easy to compute but also efficient in suggesting the ability of a model on a target dataset. 
x{In concrete, a large $l_m^{-i}$ indicates that} the $i$-th model plays a more important role in the current local task. On the contrary, when $l_m^{-i}$ is small, the $i$-th model has less effect on improving local performance. 
x{Besides, compared} with computing the loss on the whole local dataset, a batch randomly sampled at each round has the ability to describe local data attributes but has no need for massive computing resources. 

{By collecting the losses of multiple leave-one-out models, w}e construct a loss vector consisting of loss value $l_m^{-i}$ where $i \in [M]$, meaning that the performance of all participants on the current local dataset is confirmed. 
To further model the influence explicitly, we transfer the loss vector to a quantitative measurement of client-level influence. {Specifically, the definition of }client-level influence {is given as follows}.

\begin{definition}
For the $m$-th client, we define $\mathbf{\lambda}_m = [\lambda_m^1, \lambda_m^2, \cdots, \lambda_m^M] \in \mathbb{R}^{M}$ as the \textit{influence vector}. The $i$-th element in $\lambda_m$ is computed as
\begin{equation} \label{eq:inf_vec}
\lambda_m^i = \frac{[{l_m^{-i}}]^\gamma}{\sum_{i=1}^M [{l_m^{-i}}]^\gamma}
\end{equation}
\noindent where $\gamma$ is a hyper-parameter that tunes the sensitivity of clients to the influence. Setting $\gamma=0$ means that all clients share the same model aggregation weight rather than considering different levels of contribution from different clients. A larger $\gamma$ means that clients contributing more to the local performance can further strengthen their influence, thus potentially dominating the model parameters at the beginning of each communication round. 
\end{definition}

The normalization in Eq.~\eqref{eq:inf_vec} converts the leave-one-out losses into a probability-like influence distribution over all clients. A larger value of $l_m^{-i}$ means that excluding client $i$ causes a larger degradation on client $m$'s local batch, implying that client $i$ provides more useful knowledge to client $m$. Consequently, client $i$ should receive a larger aggregation weight when constructing the personalized representation model for client $m$. Conversely, if the loss remains small after removing client $i$, then the contribution of client $i$ to client $m$ is relatively limited, and its aggregation weight should be reduced.

The hyper-parameter $\gamma$ controls the sharpness of the influence distribution. When $\gamma$ is small, the resulting weights become smoother and FedC$^2$I behaves closer to uniform aggregation. When $\gamma$ is large, the aggregation becomes more selective, allowing the target client to focus on a smaller set of highly influential clients. In this sense, $\gamma$ balances knowledge sharing and personalization. A moderate value of $\gamma$ can preserve collaborative learning while avoiding excessive influence from unrelated clients.

{During the parameter aggregation phase, t}he client-level influence vector {can be} further used to build a personalized feature representation model for client $m \in [M]$ as
\begin{equation} \label{eq:p_fe}
{\bar{\theta}}_m = \sum_{i=1}^M \lambda_m^i \theta_i.
\end{equation}
\noindent Note that the influence of client $m$ on itself is also considered. By scaling up those clients with greater influence on others, a stronger collaboration between clients is achieved. 

\subsection{Class-Level Influence Measure}  \label{subsec:class_level}
By considering client-level influence, the feature representation layers are aggregated toward more optimal local performance. 
{However, class-level diversity, as an inherent characteristic of heterogeneous data, is still out of consideration. }
Usually, heterogeneous clients may not share the same decision pattern for every class. Even if two clients are globally similar, their knowledge about a specific class may still be different due to domain-specific visual patterns, class imbalance, or local annotation characteristics. For example, in digit recognition, clients may have similar representations for digits such as ``0'' and ``8'', but exhibit different decision boundaries for ambiguous digits such as ``7''. A single client-level aggregation weight is unable to capture such fine-grained differences.

To address this issue, \ourmethod evaluates the contribution of each client to each class separately. 
we apply the leave-one-out strategy to see how the performance of the classifier differs without the contribution of a class-specific weight vector. In this way, the influence from the more fine-grained level is successfully measured. 
Specifically, for the $c$-th class, we remove the class-specific classifier vector from client $i$ and observe the resulting loss on the target client's local batch. If the loss increases significantly, it indicates that the class-specific knowledge from client $i$ is beneficial for client $m$ with respect to class $c$. Therefore, the corresponding element $\Lambda_m^{i,c}$ should be assigned a larger value. This class-wise influence matrix enables the target client to construct a personalized classifier by selecting useful class-specific knowledge from different clients.

{Here, we denote} the classifier at client $m$ as $\varphi_m=[\varphi_{m,1}, \varphi_{m,2}, \cdots, \varphi_{m,C}]$, where each $\varphi_{m,c}$ in $\varphi_m$ represents the weight vector corresponding to class $c$~\cite{luo2021no}. 
{To see how the local performance differs without the contribution from the $i$-th client to the $c$-th class, following the leave-one-out principle, we first remove} the class-specific weight vector with respect to the $c$-th class and generate the weight vector by
\begin{equation} \label{eq:inf_mat}
\varphi^{-i}_{m,c}=\frac{1}{M-1} \sum_{m \in [M], m \neq i} {\varphi}_{m,c}.
\end{equation}
\noindent Then, we replace the $c$-th weight vector in $\varphi_{m}$, {\textit{i.e.}, $\varphi_{m,c}$}, by $\varphi^{-i}_{m,c}$ 
{and compute the loss $l_m^{-i,-c}$ on a random batch sampled from $\mathcal{D}_m$.}
Similar to the definition and computation of client-level influence vector, {we construct an influence matrix to indicate the class-level influence on each client, which can be defined as below.} 

\begin{definition}
For the $m$-th client, we define $\mathbf{\Lambda}_m \in \mathbb{R}^{M \times C}$ as its \textit{influence matrix}:
\begin{equation}
\mathbf{\Lambda}_m=\left\{\begin{array}{cccc}
\Lambda_m^{1,1} & \Lambda_m^{1,2} & \cdots & \Lambda_m^{1,C} \\
\Lambda_m^{2,1} & \Lambda_m^{2,2} & \cdots & \Lambda_m^{2,C} \\
\vdots & \vdots & \ddots & \vdots \\
\Lambda_m^{M,1} & \Lambda_m^{M,2} & \cdots & \Lambda_m^{M,C}
\end{array}\right\},
\end{equation}
\noindent where $M$ and $C$ represent the number of clients and the number of classes, respectively. The element at the $i$-th row and the $c$-th column, denoted as $\Lambda_m^{i,c}$, is computed by
\begin{equation}
\Lambda_m^{i,c} = \frac{[{l_m^{-i,-c}}]^\gamma}{\sum_{i=1}^M [{l_m^{-i,-c}}]^\gamma}.
\end{equation}
\noindent {$\Lambda_m^{i,c}$} represents the influence brought by the $i$-th client regarding to the $c$-th class. For the sensitivity hyper-parameter $\gamma$, we use the same value as in Eq.~\eqref{eq:inf_vec} for simplicity. 
\end{definition}

{Intuitively, similar semantic knowledge toward the same class can help clients to boost their performance when sharing their class-specific weight vector, which corresponds to a larger $\Lambda_m^{i,c}$ value. In contrast, a smaller $\Lambda_m^{i,c}$ indicates the diverse understanding toward a class, which may deteriorate the performance of local model~\cite{luo2021no}.}
{To leverage the quantified influence during the aggregation of classifier}, we use the class-level influence matrix to build the personalized local classifier $\bar{\varphi}_m=[\bar{\varphi}_{m,1}, \bar{\varphi}_{m,2}, \cdots, \bar{\varphi}_{m,C}]$, where the $c$-th element is computed by
\begin{equation} \label{eq:p_cls}
{\bar{\varphi}}_{m,c} = \sum_{i=1}^M \Lambda_m^{i,c} \varphi_{m,c}.
\end{equation}

\noindent{\textbf{Remark.}} By measuring client-level and class-level influence at the beginning of local training, a client {is able to} aggregate the feature representation layers and classifier with a unique weight allocation among participating clients. Therefore, clients {can} selectively acquire knowledge from others in a personalized but efficient way. 

\subsection{Influence-Oriented Personalized Aggregation}

After obtaining the client-level influence vector and the class-level influence matrix, FedC$^2$I performs personalized aggregation for each client. Unlike FedAvg, which produces a single global model shared by all clients, FedC$^2$I constructs a distinct aggregated model for every target client. For client $m$, the representation layers are aggregated by weighting the representation parameters of all clients with $\lambda_m$. This produces $\bar{\theta}_m$, which emphasizes the representation knowledge from clients that have a stronger influence on client $m$.

For the classifier, FedC$^2$I adopts a more fine-grained aggregation strategy. Instead of aggregating the whole classifier with one scalar weight per client, the classifier is aggregated class by class. For each class $c$, the corresponding classifier vector is obtained by combining the class-specific classifier vectors from all clients according to the $c$-th column of $\Lambda_m$. In this way, the target client can learn different classes from different sources. For example, one client may provide useful knowledge for class $c_1$, while another client may be more helpful for class $c_2$. Such a design is especially useful under feature shift, where clients may share partial semantic or visual characteristics but differ in other aspects.

The resulting personalized model $\{\bar{\theta}_m, \bar{\phi}_m\}$ serves as the initialization for subsequent local training. This design has two advantages. First, it enables each client to start local optimization from an influence-aware model that already incorporates useful cross-client knowledge. Second, it avoids forcing all clients to move toward the same global model, thereby reducing the risk of negative transfer caused by heterogeneous data distributions.

\subsection{Client Update and Optimization}  \label{subsec:update}
After defining the client-level and class-level influence measures and the corresponding aggregation schemes, we now describe how they are integrated into the complete training procedure of \ourmethod.

At each communication round, the server broadcasts the latest local parameters collected from all participating clients. Upon receiving these parameters, each client performs a two-stage local update, consisting of an influence-oriented aggregation stage and a local training stage.

In the influence-oriented aggregation stage, client $m$ first samples a random mini-batch from its private dataset and estimates the influence of other clients using only the received model parameters. Based on the estimated influences, client $m$ computes the client-level influence vector $\lambda_m$ and the class-level influence matrix $\Lambda_m$. Since this estimation is conducted locally, no raw data are transmitted to the server or other clients.

Then, client $m$ constructs an influence-oriented personalized model. Specifically, the feature representation layers are aggregated according to the client-level influence vector, as formulated in Eq.~\eqref{eq:p_fe}, while the classifier is aggregated in a more fine-grained manner using the class-level influence matrix, as formulated in Eq.~\eqref{eq:p_cls}. Through this aggregation scheme, each client obtains a customized initialization for its local model, where the feature representation parameters and classifier parameters are initialized as $\bar{\theta}_m$ and $\bar{\varphi}_m$, respectively.

In the local training stage, client $m$ further optimizes the personalized model on its private dataset using standard stochastic gradient descent or its variants. After local training is completed, the updated model parameters are sent back to the server for the next communication round.

It is worth noting that the aggregation weights are recomputed at every communication round. Therefore, the influence relationships among clients are dynamic rather than fixed. In early rounds, the influence distribution may be relatively smooth because the local models have not yet learned sufficiently discriminative knowledge. As training proceeds, the influence values gradually reflect more stable cross-client relationships, enabling each client to identify useful collaborators more effectively.
The overall training procedure of \ourmethod is summarized in Algorithm~\ref{alg:alg1}.

\noindent{\textbf{Remark.}} Influence-based measurement during the local update is beneficial in the following two aspects: \ding{182}~Instead of weighing clients with a unified criterion, the influence-oriented scheme applied here allows clients to have their unique aggregation schemes due to the heterogeneous data distributions. \ding{183}~Since the model aggregation process is moved from the server side to the client side, the server is free from performing model aggregation in each round, making it potential for the FL system to evolve from a centralized to a decentralized manner. 

\setlength{\textfloatsep}{4pt}
\begin{algorithm}[tp]
    \caption{\textbf{  \ourmethod}}
    \label{alg:algorithm}
    \textbf{Input}: $\mathcal{D}_m$, $\theta_m$, $\varphi_m$, $m=1,\cdots, M$, and $T$ \\
    \textbf{Output}: $\theta_m, \varphi_m, m=1,\cdots, M$\\
    {\bf Server executes:} \\
    \vspace{-0.4cm}
    \begin{algorithmic}[1] 
        \STATE Initialize and distribute model parameters for each client.
        \FOR{each round $t \in \{1,2,\cdots,T\}$}
            \FOR{each client $m \in \{1,2,\cdots,M\}$ in parallel} 
                \STATE $\{\theta_m, \varphi_m\} \leftarrow$ {\bf LocalUpdate}$(m, \{\theta_i\}_{i=1}^{M}, \{\varphi_i\}_{i=1}^{M})$
            \ENDFOR
        \ENDFOR     
    \end{algorithmic} 
    {\bf LocalUpdate}$(m, \{\theta_i\}_{i=1}^{M}, \{\varphi_i\}_{i=1}^{M})$:  \\
    \vspace{-0.4cm}
    \begin{algorithmic}[1] 
        \STATE Sample a batch $B$ from $\mathcal{D}_m$.
        \STATE Compute client-level influence $\mathbf{\lambda}_m$ by Eq.~(\ref{eq:inf_vec})
        \STATE Compute class-level influence $\mathbf{\Lambda}_m$ by Eq.~(\ref{eq:inf_mat})
        \STATE Build the personalized local model by Eq.~(\ref{eq:p_fe}) and Eq.~(\ref{eq:p_cls}).
        \FOR{each local epoch}
            \FOR{each batch in $D_m$}
                \STATE Update local model on private dataset $\mathcal{D}_m$
            \ENDFOR
        \ENDFOR
        \STATE \textbf{Return} local parameters $\{\theta_m, \varphi_m\}$
    \end{algorithmic}
    \label{alg:alg1}
    \setlength{\textfloatsep}{2pt}
\end{algorithm}

\subsection{Complexity and Privacy Discussion}
\textbf{Complexity Discussion}. \ourmethod introduces additional computation for influence estimation. For client-level influence, each target client evaluates $M$ leave-one-out representation models on a sampled local batch. For class-level influence, the client further evaluates class-specific leave-one-out classifiers. Since these computations are performed on a mini-batch rather than the entire local dataset, the additional cost is controlled and does not scale with the full size of the local training data. In practice, the influence estimation can be efficiently implemented by reusing forward computations and only evaluating the sampled batch before local training. 
The communication cost of \ourmethod is comparable to standard parameter-sharing FL methods. 
\ourmethod does not require transmitting raw data, data statistics, gradients on private samples, or additional public datasets. The influence vector and influence matrix can also be computed locally and do not need to be shared across clients.

\textbf{Privacy Discussion}. From a privacy perspective, \ourmethod preserves the basic privacy protocol of federated learning. All influence measurements are derived from local loss evaluations on private mini-batches, and the private data remain on the client side throughout the training process. Compared with methods that rely on explicit distribution matching or shared public data, \ourmethod estimates client relationships only through model behavior, making it suitable for privacy-sensitive heterogeneous FL scenarios.

\begin{table*}[t!]
    \caption{Test accuracy of different FL methods on Digit-5 which includes five datasets containing ten overlapping categories. The results are in mean (std) format over three independent runs with different random seeds.}
    \label{tab:perf_digit}
    \centering
    \begin{centering}
    \begin{tabular}{p{3cm}|p{1.9cm}<{\centering} p{1.9cm}<{\centering} p{1.9cm}<{\centering} p{1.9cm}<{\centering} p{1.9cm}<{\centering}| p{1.9cm}<{\centering}}
    \toprule
    \textbf{Method} & MNIST & SVHN & USPS & SynthDigits & MNIST-M & \textbf{Avg} \\
    \midrule
    Local & 94.02(0.44) & 56.63(2.66) &	93.73(1.48) & 74.95(0.83) & 79.72(1.24) & 79.81(0.60) \\
    \midrule
    FedAvg~\cite{mcmahan2016communication} &  95.45(0.52) & 68.17(0.95) & 93.22(0.89) & 80.73(0.49) & 76.82(1.08) & 82.88(0.04) \\
    FedProx~\cite{li2018federated} & \textbf{95.60}(0.18) & 68.12(1.72) & 93.63(1.18) & 80.53(0.38) & 76.77(1.07) & 82.93(0.15) \\
    \midrule
    FedRep~\cite{collins2021exploiting} & 95.18(0.93) & 68.70(2.13) & 94.13(1.81) & {81.07}(2.92) & 77.12(0.63) & 83.24(0.93) \\
    FedRoD~\cite{chen2022on} & {94.97}(0.21) & 69.45(1.03) & 94.60(1.75) & \textbf{82.07}(0.66) & 78.47(1.65) & 83.92(1.39) \\
    FedProto~\cite{tan2021fedproto} & 95.35(0.87) & 69.33(0.84) & 94.99(1.64) & 77.88(0.97) & 78.27(0.65) & 83.16(1.20) \\
    FedCALM~\cite{zheng2025fedcalm} & 94.76(0.65) & 70.67(1.03) & 94.80(1.44) & 80.95(2.31) & 78.57(1.13) & 83.95(1.31)\\
    FedIC~\cite{zheng2026intra} & 94.06(0.96) & 68.24(1.48) & 94.52(1.29) & 79.04(2.05) & 79.19(1.06) & 83.01(1.37)\\
    \midrule
    Ours & 95.00(0.93) & \textbf{72.78}(0.99) & \textbf{95.08}(0.35) & 81.05(1.93) & \textbf{83.60}(0.94) & \textbf{85.50}(0.63) \\
    \bottomrule
    \end{tabular}
    \end{centering}    
\end{table*}

\begin{table*}[htbp!]
    \caption{Test accuracy of different FL methods on Office-10 which includes four datasets containing ten overlapping categories. The results are in mean (std) format over three independent runs with different random seeds.}
    \label{tab:perf_office}
    \centering
    \begin{centering}
    \begin{tabular}{p{3cm}|p{1.9cm}<{\centering} p{1.9cm}<{\centering} p{1.9cm}<{\centering} p{1.9cm}<{\centering} |p{1.9cm}<{\centering}}
    \toprule
    \textbf{Method} & Amazon & Caltech & DSLR & WebCam & \textbf{Avg} \\
    \midrule
    Local &  60.24(0.60) & 28.15(1.56) & 76.04(1.80) & 76.84(2.59) & 60.32(0.09) \\
    \midrule
    FedAvg~\cite{mcmahan2016communication} &  58.68(1.08) & 38.90(2.28) & 77.08(4.77) & 84.75(3.39) & 64.87(2.49) \\
    FedProx~\cite{li2018federated} &  56.77(1.56) & \textbf{38.96}(1.80) & 78.12(3.12) & \textbf{86.44}(2.94) & 65.07(0.64) \\
    \midrule
    FedRep~\cite{collins2021exploiting} &  59.67(1.70) & 38.89(1.95) & 79.12(2.41) & 82.71(2.74) & 65.10(1.80) \\
    FedRoD~\cite{chen2022on} &  60.07(0.60) & 38.15(1.03) & 79.50(2.45) & 83.17(1.17) & 65.22(1.25) \\
    FedProto~\cite{tan2021fedproto} & 61.76(1.50) & 38.25(0.93) & 78.00(2.27) & 82.27(0.87) & 65.07(1.64) \\
    FedCALM~\cite{zheng2025fedcalm} & 61.35(1.20)  & 37.78(1.20) & 79.95(2.88) & 82.5(2.25) & 65.40(1.88)\\
    FedIC~\cite{zheng2026intra} &60.12(0.74) & 37.32(1.67) & 78.54(3.04) & 82.92(3.38) & 64.73(2.21)  \\
    \midrule
    Ours & \textbf{63.72}(0.77) & 38.22(1.54) & \textbf{81.25}(2.25) & 83.62(1.53) & \textbf{66.70}(0.91) \\
    \bottomrule
    \end{tabular}
    \end{centering}
\end{table*}

\section{Experiments}

\subsection{Experimental Setup}
\paragraph{Datasets} 
We evaluate \ourmethod on two benchmark heterogeneous FL settings: Digit-5~\cite{zhou2020learning} and Office-10~\cite{gong2012geodesic}. The former contains ten digit classes from five domains, namely MNIST, SVHN, USPS, SynthDigits, and MNIST-M. The latter contains ten overlapping categories from four domains, including Amazon, Caltech, DSLR, and WebCam. To simulate the data heterogeneous scenario in FL, each dataset owned by a client is from a different domain. 

\paragraph{Visualization of Raw Samples}
To provide an intuitive understanding of the data heterogeneity across different clients, we visualize several raw samples from Digit-5 and Office-10 in Figure~\ref{fig:raw_instance}. 
For each dataset, we present representative samples from five classes across its sub-datasets. 
As observed, although samples from different sub-datasets share the same semantic labels, they exhibit clear visual discrepancies in style, background, color distribution, and image quality. 
Such feature shifts indicate the existence of non-identical data distributions among clients, which further motivates the need for personalized aggregation in federated learning.

\begin{figure}[t!]
	\centering
	\subfigure[Digit-5]{
		\begin{minipage}[b]{0.226\textwidth}
			\includegraphics[width=1\textwidth]{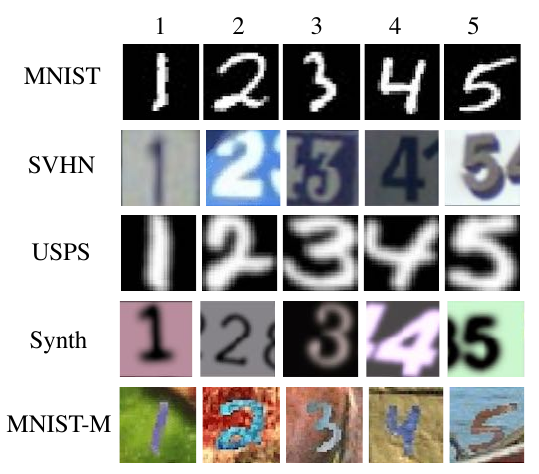}
		\end{minipage}
		\label{fig:digit_raw}
	}
	\subfigure[Office-10]{
		\begin{minipage}[b]{0.226\textwidth}
			\includegraphics[width=1\textwidth]{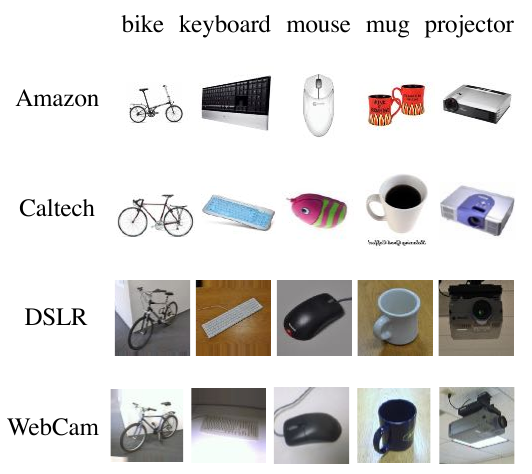}
		\end{minipage}
		\label{fig:office_raw}
	}
	\caption{Examples of raw instances from two datasets: Digit-5 (\textit{left}) and Office-10 (\textit{right}). We present five classes for each dataset to show the feature shift across their sub-datasets.}
    \label{fig:raw_instance}
\end{figure}

\paragraph{Baselines} 
We compare \textit{\ourmethod} with eight baselines, including the pure local training algorithm, vanilla FL methods that tackle non-IID problems in FL, and state-of-the-art personalized FL methods. The details of these baselines are provided as follows.
\begin{itemize}
    \item \textbf{Local}: Each client trains their local model based on the local data without communication with others. 
    \item \textbf{FedAvg}~\cite{mcmahan2016communication}: Clients send learnable parameters to the server and receive aggregated parameters from the server.
    \item \textbf{FedProx}~\cite{li2018federated}: Based on FedAvg, a regularization term with importance weight $\mu$ is added to the original loss function. In our experiments, $\mu$ is set to 0.01.
    \item \textbf{FedRep}~\cite{collins2021exploiting}: Compared with FedAvg, FedRep decouples the model into two parts, \textit{i.e.}, globally shared representation layers and personalized client-specific heads. 
    \item \textbf{FedRoD}~\cite{chen2022on}: FedRoD decouples the duty of the local model for generic FL and personalized FL, respectively. 
    \item \textbf{FedProto}~\cite{tan2021fedproto}: Class-specific prototypes, instead of model parameters serve as the information carrier in FL. 
    \item \textbf{FedCALM}~\cite{zheng2025fedcalm}: FedCALM mitigates server-side aggregation conflicts through a layer-wise aggregation strategy. 
    \item \textbf{FedIC}~\cite{zheng2026intra}: 
    FedIC performs intra-class prototype clustering and enables classifier collaboration within clusters.
\end{itemize}

\paragraph{Implementation Details}
We use the LeNet~\cite{lecun1998gradient} as the local model, where there are two convolutional layers with a kernel size of 5. 
The local epoch number and batch size are 2 and 32, respectively.  
The number of communication rounds is 20 and 40 for Digit-5 and Office-10, respectively. We report the results with the average over 3 runs of different random seeds. All the methods are implemented by PyTorch, and all experiments are conducted on one NVIDIA RTX 4090 GPU. 

We use an Adam~\cite{kingma2014adam} optimizer with weight decay of 0 and learning rate of 0.001 by grid search from $\{7e-4,5e-4,3e-4,1e-4,0\}$ and $\{5e-3, 1e-3, 5e-4, 1e-4\}$, respectively.
For the hyper-parameter $\gamma$, we use the best value $\gamma=5$ for both Digit-5 and Office-10 found by grid search from $\{0.5, 1, 2, 5, 10\}$. 
The grid search for the optimal value of these sensitive hyper-parameters is conducted on the validation dataset, while the rest insensitive hyper-parameters are fixed. 

\begin{figure*}[h]
    \centering
    \subfigure[Round 1]{
        \begin{minipage}[b]{0.23\textwidth}
            \includegraphics[width=1\textwidth]{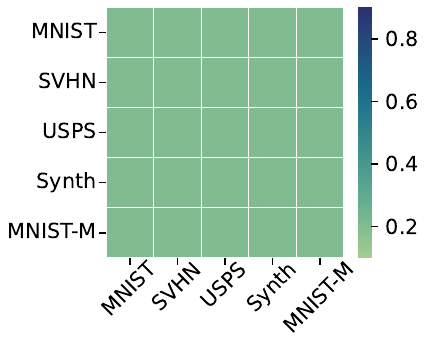}
        \end{minipage}
    }
    \subfigure[Round 2]{
        \begin{minipage}[b]{0.23\textwidth}
            \includegraphics[width=1\textwidth]{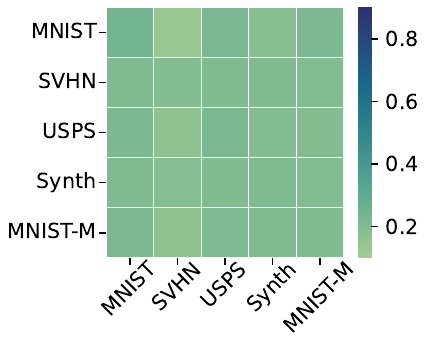}
        \end{minipage}
    }
    \subfigure[Round 10]{
        \begin{minipage}[b]{0.23\textwidth}
            \includegraphics[width=1\textwidth]{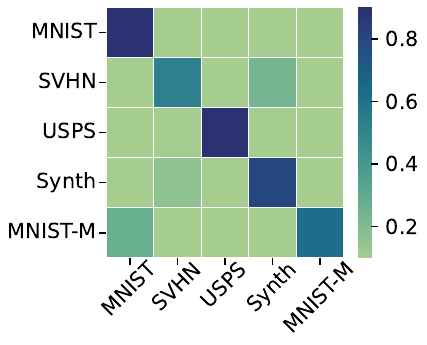}
        \end{minipage}
    }
    \subfigure[Round 20]{
        \begin{minipage}[b]{0.23\textwidth}
            \includegraphics[width=1\textwidth]{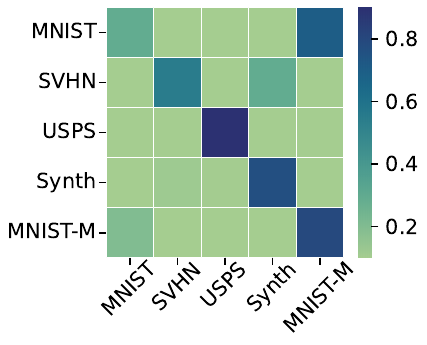}
        \end{minipage}
    }
    \vspace{0.1cm}
    \subfigure[Round 1]{
        \begin{minipage}[b]{0.23\textwidth}
            \includegraphics[width=1\textwidth]{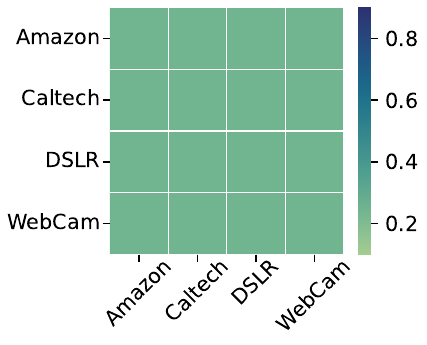}
        \end{minipage}
    }	
    \subfigure[Round 4]{
        \begin{minipage}[b]{0.23\textwidth}
            \includegraphics[width=1\textwidth]{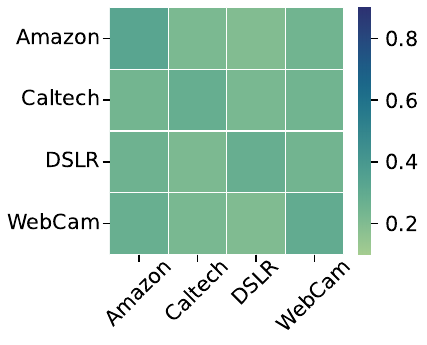}
        \end{minipage}
    }
    \subfigure[Round 10]{
        \begin{minipage}[b]{0.23\textwidth}
            \includegraphics[width=1\textwidth]{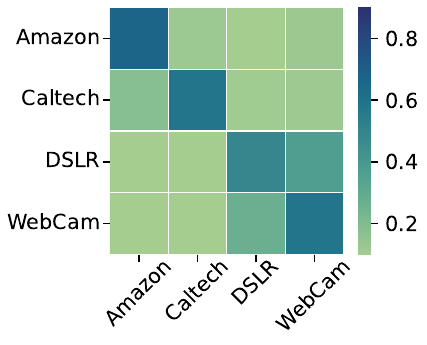}
        \end{minipage}
    }
    \subfigure[Round 40]{
        \begin{minipage}[b]{0.23\textwidth}
            \includegraphics[width=1\textwidth]{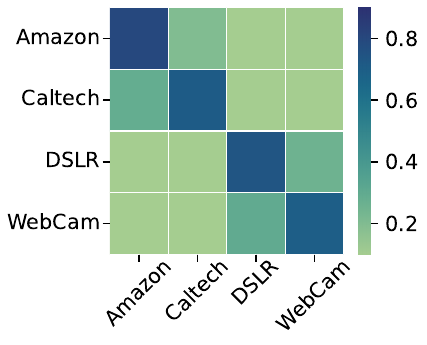}
        \end{minipage}
    }
    \caption{Visualization of influence vector in different communication rounds. (a)-(d) The value of the influence vector in round 1, 2, 10, and 20 on Digit-5. (e)-(h) The value of the influence vector in round 1, 4, 10, and 40 on Office-10. Each figure shows the client-level influence values over all clients. }
    \label{fig:vis_iv}
    \vspace{-4mm}
\end{figure*}

\subsection{Performance Comparison} 
We compare our proposed \ourmethod {with} the baseline methods to verify that {effectiveness of} \ourmethod in feature shift non-IID {scenarios}. Table~\ref{tab:perf_digit} and~\ref{tab:perf_office} report the results on Digit-5 and Office-10, respectively. The results are in mean (std) format over three independent runs with different random seeds. 

From the results, we can see that \ourmethod outperforms the best baseline method in at least half of the clients/domains and achieves a certain improvement by $0.78\%-2.62\%$ on the average test accuracy. 
Moreover, it is worthwhile to note that there is a significant performance boost in MNIST-M. {The potential reason is that} some similar domain properties shared by MNIST-M and other clients, \textit{i.e.}, MNIST, are recognized and {leveraged by \ourmethod} during the parameter aggregation procedure in the format of influence measurement.

\subsection{Ablation Studies} 
To verify the effect of the proposed influence vector and influence matrix that are used to measure client-level and class-level influence, we conduct an ablation study by comparing \ourmethod with its variants. {Specifically, we consider} three variants including \ding{182}~only client-level influence is measured (\ourmethod\ - $\lambda$); 
\ding{183}~only class-level influence is measured while the feature representation $\theta$ is locally trained (\ourmethod\ - $\Lambda$/\textit{l}); 
\ding{184}~only class-level influence is measured while the feature representation $\theta$ is globally shared following vanilla parameter aggregation scheme (\ourmethod\ - $\Lambda$/\textit{g}). 
As shown in Table~\ref{tab:ablation}, client-level influence measurement directly contributes to the final performance improvement, while class-level influence makes its contribution based on the client-level influence. 

\begin{table}
    \caption{Ablation studies on the effects of influence vector $\lambda$ and influence matrix $\Lambda$.}
    \label{tab:ablation}
    \centering
    \begin{tabular}{p{2cm}|p{2.cm}<{\centering} p{2.cm}<{\centering}}
        \toprule
        \textbf{Method}  & Digit-5 & Office-10 \\
        \midrule
        Local     & 79.81(0.60) & 60.32(0.09) \\
        FedAvg     &  82.88(0.04) & 64.87(2.49) \\
        \midrule
        \ourmethod\ - $\lambda$  & 84.61(0.24) & 63.07(2.85) \\
        \ourmethod\ - $\Lambda$/\textit{l} & 79.32(2.58) & 62.53(0.76) \\
        \ourmethod\ - $\Lambda$/\textit{g} & 84.85(0.34) & 64.08(2.92) \\
        \midrule
        \ourmethod  & \textbf{85.50}(0.63) & \textbf{66.70}(0.91) \\
        \bottomrule
    \end{tabular}
\end{table}

\subsection{Effects of Varying $\gamma$} 
The hyper-parameter $\gamma$ plays an important role in tuning the sensitivity of clients to influences. A larger $\gamma$ makes the system more influence-sensitive and strengthens the influence of those important clients. We tune $\gamma$ from the candidate set $\{0.5, 1, 2, 5, 10\}$ and select the one with the best performance on the validation dataset. In Table~\ref{tab:gamma}, we provide the results of \ourmethod with varying values of $\gamma$ under Digit-5 and Office-10 datasets. {As we can see, t}he best value of $\gamma$ is 5 {for both datasets}. There can be reasonable techniques developed for adaptively choosing appropriate $\gamma$, which can be further discovered in future works.\looseness-1

\begin{table}
    \caption{Effect of varying $\gamma$.}
    \label{tab:gamma}
    \centering
    \begin{tabular}{p{1.cm}|p{2.5cm}<{\centering} p{2.5cm}<{\centering}}
        \toprule
        \multirow{2}*{\makecell[l]{$\gamma$}}  & \multicolumn{2}{c}{Dataset}\\
        \cmidrule{2-3}
        & Digit-5  & Office-10 \\
        \midrule
        0.5 & 85.03(0.11) & 63.07(2.76) \\
        1 &  85.08(0.60) & 64.20(1.12) \\
        2 & 85.48(0.27) & 64.25(2.11) \\
        5 & \textbf{85.50}(0.63) & \textbf{66.70}(0.91) \\
        10 & 84.26(0.55) & 66.18(2.92) \\
        \bottomrule
    \end{tabular}
\end{table}

\subsection{Visualization of Client-Level Influence}
To better understand how the influence measurement is carried out and how the client-level influence varies in different communication rounds, we visualize the values of influence vector in Fig.~\ref{fig:vis_iv}. The figures in the top row are for Digit-5, while the bottom row is for Office-10. It can be seen that the influence from different clients becomes slightly different in the first several rounds. Then, clients are stably influenced by themselves and some of other clients who share similar inherent data properties. For example, in the Digit-5 experimental setting, clients owning SVHN and SynthDigits have similar font styles, so the results show that they are more likely to be influenced by each other compared with other clients. 
Also, it is worthwhile to notice that the visualized client-level influence is not always symmetric in the client-based coordinates, making it a more general client-wise instead of a pair-wise measurement. The reason for that is we measure the influence in a unidirectional way at the local side rather than measuring the similarity between two clients in a bidirectional way at the central side. 

\section{Conclusion}
In this paper, we propose a novel federated learning framework, namely \ourmethod, that measures both client-level and class-level influence to achieve a more efficient model aggregation and benefit more participating clients. We construct the influence vector to specify the influence brought by other clients and {leverage the client-level influence to} aggregate the feature representation layers. We also construct the influence matrix to aggregate class-level knowledge in a more fine-grained manner, {which enables each client to learn a class-personalized local classifier}. Experimental results illustrate the superiority of \ourmethod compared with baselines and verify the effect of the key components in \ourmethod.